\documentclass[10pt,twocolumn,letterpaper]{article}

\usepackage{cvpr}
\usepackage{times}
\usepackage{epsfig}
\usepackage{graphicx}
\usepackage{amsmath}
\usepackage{amssymb}

\usepackage{algorithm, algorithmic, multirow, amsmath, xcolor}
\usepackage{caption,subfigure}
\usepackage[pagebackref=true,breaklinks=true,letterpaper=true,colorlinks,bookmarks=false]{hyperref}

\cvprfinalcopy 

\def\cvprPaperID{1741} 

\ifcvprfinal\fi
\begin{document}

\title{B-spline Shape from Motion \& Shading: An Automatic Free-form Surface Modeling for Face Reconstruction}

\author{Weilong Peng\\
School of Computer Science, Tianjin University\\
{\tt\small wlpeng@tju.edu.cn}
\and
Zhiyong Feng\\
School of Computer Science, Tianjin University\\
{\tt\small zyfeng@tju.edu.cn}
\and
Chao Xu\\
School of Software, Tianjin University\\
Tianjin, 300072, China\\
{\tt\small xuchao@tju.edu.cn}}

\maketitle

\begin{abstract}
Recently, many methods have been proposed for face reconstruction from multiple images, most of which involve fundamental principles of Shape from Shading and Structure from motion. However, a majority of the methods just generate discrete surface model of face. In this paper, B-spline Shape from Motion and Shading (BsSfMS) is proposed to reconstruct continuous B-spline surface for multi-view face images, according to an assumption that shading and motion information in the images contain 1st- and 0th-order derivative of B-spline face respectively. Face surface is expressed as a B-spline surface that can be reconstructed by optimizing B-spline control points. Therefore, normals and 3D feature points computed from shading and motion of images respectively are used as the 1st- and 0th- order derivative information, to be jointly applied in optimizing the B-spline face. Additionally, an IMLS (iterative multi-least-square) algorithm is proposed to handle the difficult control point optimization. Furthermore, synthetic samples and LFW dataset are introduced and conducted to verify the proposed approach, and the experimental results demonstrate the effectiveness with different poses, illuminations, expressions etc., even with wild images.
\end{abstract}

\section{Introduction}

3D face has been widely used in areas of face recognition and verification, expression recognition and facial animations. Face reconstruction is usually based on a scene that the subject is fixed and scanned with special equipment under a lab condition, video sequences or multi-view photographs. Recently, more challenging works are reconstruction from wild images with unknown camera calibration.

As a whole, there are two kinds of reconstruction technologies: high-cost equipment relied technologies under lab condition and low-cost 2D image based techniques out of lab. The former ones have implemented high quality face reconstruction such as laser~\cite{6}, stereo~\cite{4,8,9}, structural lighting~\cite{3}, and light stages~\cite{12}. These technologies always rely on high resolution camera, synchronized data or calibration. The latter kinds of reconstructions are based on 2D informations, such as single-image method~\cite{27}, video sequences based method~\cite{1}, multi-view photographs~\cite{2,23}, and even wild images based methods~\cite{30,40}. In contrast, the latter are low-cost and convenient to be distributed. It is greatly significant but challengingly to research on face reconstruction from images. 

Generally, 2D methods involves several kinds of fundamental methods including Shape from Shading (SFS)~\cite{18}, Structure from Motion(SFM)~\cite{26}, and 3D Morphable Model (3DMM)~\cite{35,34}. SFS assumes that the subject has a Lambertian surface and relatively distant illumination, to compute surface normals. SFM is based on assumption that it exists a coordinate transformation between image coordinate system and camera coordinate system, to compute the positions of surface points. And the idea of 3DMM is that human faces are within a linear subspace, and that any novel face shape can be represented by linear combination of shape eigenvectors deduced by PCA. SFS and SFM give the geometrical and physical descriptions of face shape and imaging, and 3DMM concentrates on the statistical explanation of 3D meshes or skeletons. We believe face reconstruction is rather a geometrical optimization problem than a statistical problem, as 3DMM is just an assist of geometrical method when building detailed shape,eg.\cite{5}. 

Motivated by the geometric and physical thought in SFS and SFM, we propose to use B-spline surface to represent face shape, because it is a continual surface model, while 3D mesh or skeleton model are just approximation of continuous face by using discontinuous fragments. Deep in relation between B-spline face and face images, we find that motion and shading in images are from 0th- and 1st- order derivatives, respectively, of B-spline face surface, by which a complete geometrical optimization for B-spline surface reconstruction can be introduced: 1) the dense shading information can be applied to optimize local B-spline shape; and 2) the sparse motion can be used to optimize the global. During optimization, 2nd-order informations are also used to guarantee a smooth warping and local information keeping. The processes 1) and 2) iterate until convergence. Significantly, the final B-spline face is high-order differential, and can approximate the true shape in arbitrary precision.

\noindent \textbf{Shape in Shading and Structure in Motion.} SFS has been widely used for reconstruction, e.g. single-view reconstruction~\cite{13}, multiple frontal images based reconstruction~\cite{44}, and wild image based reconstruction ~\cite{30,40}. As single-view is ill posed~\cite{19}, a reference is always needed~\cite{13}. For wild images, photometric stereo is applied to obtain accurate normals locally~\cite{30,40}.  SFM uses multiple frame or images to recover sparse 3D structure of feature points of an object~\cite{26}. Spatial-transformation approach ~\cite{29} only estimates the depth of facial points. Bundle adjustment~\cite{24} fits the large scale rigid object reconstruction, but it cannot generate the dense model of non-rigid face. Incremental SFM approach~\cite{28} is proposed to build a 3D generic face model for non-rigid face. The paper~\cite{40} optimizes the local information with normals from shading, based on a 3D feature points-driven global warping.  Therefore, shading and motion are very important and different geometric information of face, will enhance the reconstruction when they are combined. In our method, they are integrated into B-spline face model as 1st- and 0th- order derivatives of spline surface.

\noindent \textbf{Face Surface Modeling.} The surface modeling depends on the data input (point cloud, noise, outlier, etc), output (point cloud, mesh, skeleton), and types of shape(man-made shape, organic shape). Point cloud, skeleton and mesh grid are the widely used man-made shape type for face reconstruction. Spatial transformation method~\cite{29} estimates positions of sparse facial feature points. Bundle adjustment~\cite{24} builds the dense point cloud for large scale rigid object with a great number of images. Jingu et al ~\cite{23} reconstruct face dense mesh based on skeleton and 3DMM. Kemelmacher et al~\cite{30} apply integration of normals to get discrete surface points. Roth et al~\cite{30} reconstruct face based on Laplace mesh editing. The method of surface-smoothness priors is also needed for reconstructing smooth mesh on point cloud, e.g. radial basis function~\cite{31} and Poisson surface reconstruction~\cite{43}. Due to the fact that point cloud and 3D mess are discontinuous geometric shape, they cannot approximate the true shape of face in arbitrary precision. In contrast, our technology is based on a continuous free-form surface that approximates a organic shape, and reconstruct B-spline face from images directly, instead of intermediate point data. What's more, because of B-spline surface is a case of NURBS (Non-Uniform Rational B-Spline), the B-spline reconstruction also can be imported to 3D modeling software like Rhino3D for further editing, analysis, and transformation conveniently by adjusting the B-spline control points.

In our work, initial local normals and motion estimation are obtained from images by referring to the work~\cite{30}, and they are updated during B-spline optimization. The B-spline face is optimized and updated locally by using normals and globally by 3D feature points respectively, and the process iterates until convergence.

In summary, there are two primary contributions:

1. The face is described as a B-spline face whose 1st- and 0th- order derivatives can be discovered from shading and motion informations respectively in the images; and iterative multi-least-square (IMLS) optimization algorithm is also given.

2. We introduce an automatic free-form surface modeling for face reconstruction from multiple images with variation such as different poses, illuminations, and expressions etc., even from images in the wild. 

\section{Methodology}
The proposed method is operated on a set of face images with different poses, illuminations, and expression, of an individual. In this section, face is modeled as a uniform B-spline surface, whose 1st- and 0th- order derivatives have intimate relations with normals in shading and 3D feature points from motion respectively in the images in Section \ref{sec:2.1}. Therefore, local and global B-spline optimization by shading and motion are formulated in Section \ref{sec:2.2} and \ref{sec:2.3} severally. Finally, an IMLS optimization algorithm is given in Section \ref{sec:2.4}.

\subsection{Modeling B-spline Face}
\label{sec:2.1}
Assuming human face is a uniform B-spline surface, surface curvature and normal at each point are described by basis function and control points, and face can be reconstructed conveniently by computing the B-spline control points. 

\begin{figure}[t]
\begin{center}
   \includegraphics[width=1.0\linewidth]{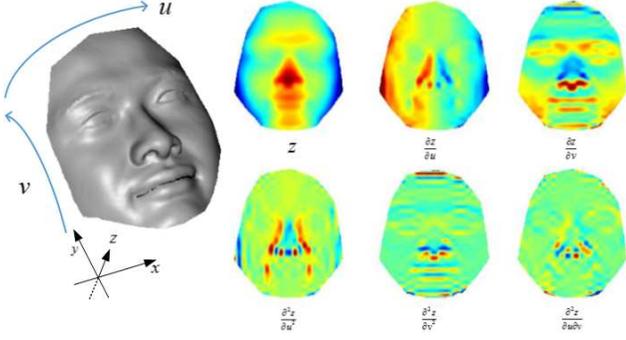}
\end{center}
   \caption{$C^2$ continuous characteristics of a $4\times4$ order B-spline face: maps of component z, $\frac{{\partial z}}{{\partial u}}$, $\frac{{\partial z}}{{\partial v}}$, $\frac{{\partial ^2 z}}{{\partial u^2 }}$, $\frac{{\partial ^2 z}}{{\partial v^2 }}$, and $\frac{{\partial ^2 z}}{{\partial u }{\partial v}}$ of the face are shown.
}
\label{fig:meth_deriv}
\end{figure}

We model face as a uniform B-spline surface $\textbf{\textit{F}}$ of degree $4\times4$, with $\emph{\textbf{B}}=\{\emph{\textbf{b}}_{mn} \}_{M\times N}$ as the control points, and $U=\{u_m \}_{m=1}^{M+4},V=\{v_n \}_{n=1}^{N+4}$ as knots spliting $uv$ parameter plane in $u$ and $v$ direction. Then

\begin{equation}
 \textbf{\textit{F}}(u,v) = \sum\limits_{m = 1}^M {\sum\limits_{n = 1}^N {R_{m,n} (u,v)\emph{\textbf{b}}_{mn} } } ,
 \label{equ:01}
\end{equation}
with $R_{m,n}(u,v) = N_{m,4} (u) \cdot N_{n,4} (v)$ and
\[\left\{ {\begin{array}{*{20}{c}}
{{N_{i,1}}(u) = \left\{ {\begin{array}{*{20}{l}}
{\begin{array}{*{20}{c}}
1&{{u_i} \le u < {u_{i + 1}},}
\end{array}}\\
{\begin{array}{*{20}{c}}
0&{otherwise,}
\end{array}}
\end{array}} \right.}\ \ \ \ \ \ \ \ \ \ \ \ \ \ \\
{{N_{i,4}}(u) = \frac{{(u - {u_i})\cdot{N_{i,3}}(u)}}{{{u_{i + 3}} - {u_i}}} + \frac{{({u_{i + 4}} - u)\cdot{N_{i + 1,3}}(u)}}{{{u_{i + 4}} - {u_{i + 1}}}}}.
\end{array}} \right.\]
Particularly, $\textbf{\textit{F}}(u,v)$ is $C^2$-continuous, and  z-component of a B-spline face surface and its high order partial derivatives w.r.t. $u$ and $v$ are shown by the maps in Fig.\ref{fig:meth_deriv}. The continuous derivability means that $\textbf{\textit{F}}$ can approximate the true shape in arbitrary $uv$ precision with deterministic $k$-ordered partial derivative $\frac{{{\partial ^k}\textbf{\textit{F}}}}{{\partial {u^k}}}$ and $\frac{{{\partial ^k}\textbf{\textit{F}}}}{{\partial {v^k}}}$, $k=1,2$, and $\frac{{{\partial ^2}\textbf{\textit{F}}}}{\partial u\partial v}$.

\subsubsection{Surface normals from 1st-order derivative}
The 1st-order partial derivatives of $\textbf{\textit{F}}$ w.r.t $u$ and $v$ are
\begin{scriptsize}
\[
\begin{array}{l}
 \textbf{\textit{F}}_u '(u,v) = \sum\limits_{m = 1}^M {\sum\limits_{n = 1}^N {N_{m,4} '(u) \cdot N_{n,4} (v)\textbf{\textit{b}}_{mn} } }  \\
  = \sum\limits_{m = 1}^M {\sum\limits_{n = 1}^N {(\frac{4}{{u_{m + 4}  - u_i }}N_{m,3} '(u) - \frac{4}{{u_{m + 5}  - u_{m + 1} }}N_{m + 1,3} '(u)) \cdot N_{n,4} (v)\textbf{b}_{mn} } }  \\
 \end{array}
\]
\end{scriptsize}
and
\begin{scriptsize}
\[
\begin{array}{l}
 \textbf{\textit{F}}_v '(u,v) = \sum\limits_{m = 1}^M {\sum\limits_{n = 1}^N {N_{m,4} (u) \cdot N_{n,4} '(v)\textbf{\textit{b}}_{mn} } }  \\
  = \sum\limits_{m = 1}^M {\sum\limits_{n = 1}^N {N_{m,4} (u) \cdot (\frac{4}{{u_{n + 4}  - u_n }}N_{n,3} '(v) - \frac{4}{{u_{n + 5}  - u_{n + 1} }}N_{n + 1,3} '(v))\textbf{b}_{mn} } }  \\
 \end{array}
\]
\end{scriptsize}
respectively. And a more brief form of 1st-order derivatives at $(u,v)$ can be written linearly as
\begin{equation}
\left\{ {\begin{array}{*{20}{c}}
{\textbf{\textit{F}}{'_u}{|_{u,v}} = {\textbf{\textit{T}}_1}{|_{u,v}} \cdot \textbf{\textit{b}}}\\
{\textbf{\textit{F}}{'_v}{|_{u,v}} = {\textbf{\textit{T}}_2}{|_{u,v}} \cdot \textbf{\textit{b}}}
\end{array}},\right.
\label{equ:02}
\end{equation}
where vector $\textbf{\emph{b}}_{3MN\times1}$ represents B-spline control points, and $\textbf{\emph{T}}_1 |_{u,v}$ and $\textbf{\emph{T}}_2 |_{u,v}$ are 3-by-$3MN$ matrixes stacking the 1st-order coefficient of control points, being referred as 1st-order coefficient matrixes.

Therefore, the surface normal vector at $(u,v)$ can be computed by the cross product
\begin{equation}
\emph{\textbf{H}}|_{u,v}  = \frac{{\textbf{\emph{F}}_u '|_{u,v}  \times \textbf{\emph{F}}_v '|_{u,v} }}{{\left\| {\textbf{\emph{F}}_u '|_{u,v}  \times \textbf{\emph{F}}_v '|_{u,v} } \right\|}} = s|_{u,v}  \cdot \textbf{\emph{F}}_u '|_{u,v}  \times \textbf{\emph{F}}_v '|_{u,v}
\label{equ:03}
\end{equation}
\subsubsection{3D structure by 0th-order derivative}

\begin{figure}
\begin{center}
   \includegraphics[width=1.0\linewidth]{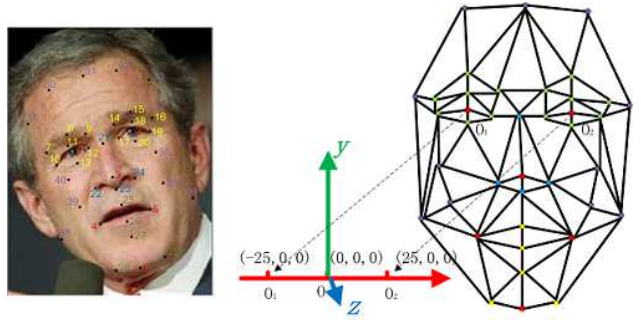}
\end{center}
   \caption{ Face structure defined by 40 feature points: the left side shows the point positions in a face image; the right side shows the structure topology with eye center points of $O_1$ (-25, 0, 0) and $O_2$ (25, 0, 0) in 3$D$ space, which looks like a frontal 2D face structure from the direction of normal (0, 0, 1).}
\label{fig:meth_structure}
\end{figure}

Particularly, 3D structure defined by $f$ feature points on the surface, as shown in Fig.\ref{fig:meth_structure}, is given intuitively as

\begin{equation}
\begin{array}{*{20}c}

F(\emph{\textbf{B}},u(i),v(i)) = \ \ \ \ \ \ \ \ \ \ \ \ \ \ \ \ \ \ \ \ \ \ \ \ \ \ \ \ \\
\begin{array}{c}
\left[ {\begin{array}{*{20}c}
   {N_{m - 3,3} (u(i))}  \\
   {N_{m - 2,3} (u(i))}  \\
    {N_{m - 1,3} (u(i))}    \\
   {N_{m,3} (u(i))}  \\
\end{array}} \right]^T \cdot \emph{\textbf{B}}_{m,n}^{(4,4)}
 \cdot \left[ {\begin{array}{*{20}c}
   {N_{n - 3,3} (v(i))}  \\
   {N_{n - 2,3} (v(i))}  \\
    {N_{n - 1,3} (v(i))}   \\
   {N_{n,3} (v(i))}  \\
\end{array}} \right], \\
\end{array} \\

\end{array}
\label{equ:04}
\end{equation}
$i = 1,2,...,f,$ and
 \[
 \emph{\textbf{B}}_{m,n}^{(4,4)}  = \left[ {\begin{array}{*{20}c}
   {\textbf{\emph{b}}_{m - 3,n - 3} } & {\textbf{\emph{b}}_{m - 3,n - 2} } &  
   {\textbf{\emph{b}}_{m - 3,n - 1} }   & 
   {\textbf{\emph{b}}_{m - 3,n} }  \\
   {\textbf{\emph{b}}_{m - 2,n - 3} } & {\textbf{\emph{b}}_{m - 2,n - 2} } &  
   {\textbf{\emph{b}}_{m - 2,n - 1} }  & 
   {\textbf{\emph{b}}_{m - 2,n} }  \\
    {\textbf{\emph{b}}_{m - 1,n - 3} }   &
      {\textbf{\emph{b}}_{m - 1,n - 2} }  &
       {\textbf{\emph{b}}_{m - 1,n - 1} } &
         {\textbf{\emph{b}}_{m - 1,n} }   \\
   {\textbf{\emph{b}}_{m,n - 3} } & {\textbf{\emph{b}}_{m,n - 2} } &  
  {\textbf{\emph{b}}_{m ,n - 1} }  & 
   {\textbf{\emph{b}}_{m,n} }  \\
\end{array}} \right].
\]

Equ.\ref{equ:04} can be rewritten linearly as
\begin{equation}
\textbf{\emph{F}}|_{u(i),v(i)}  = \textbf{\emph{T}}|_{u(i),v(i)}  \cdot \textbf{\emph{b}}, i = 1,2,...,f,
\label{equ:05}
\end{equation}
where $\textbf{\emph{T}}|_{u(i),v(i)}$ is a 3-by-$3MN$ 0th-order coefficient matrix. $\textbf{\emph{T}}|_{u(i),v(i)}$ is also a sparse matrix as every feature point is determined by only 16 control points according to  $\emph{\textbf{B}}_{m,n}^{(4,4)}$.

Surface normals are the geometric details of a surface, and structure is a sparse but global shape of a face. They are 1st- and 0th- order information that can be represented by control parameter $\textbf{\emph{b}}$, seen in Equ.\ref{equ:02}, Equ.\ref{equ:03} and Equ.\ref{equ:05}. Conversely, B-spline parameter $\textbf{\emph{b}}$ can be rebuilt from normals and structure, to build the face surface of an individual. 

\subsection{Local B-spline Shape From Shading}
\label{sec:2.2}
 The shading in images covers normals and albedo of a surface, which can be used to optimize B-spline control points of a B-spline face. Firstly, photometric stereo method is used to estimate the normal information by assuming that face images are captured from a B-spline shape by a camera model of scaled orthographic projection, relatively distant illumination and Lambertian reflectance. Secondly, a normals driven control points optimization is proposed to optimize the local shape of B-spline face.

\subsubsection{Photometric stereo}
Photometric stereo is also based on assumption that a set of images of a same person are captured under different illumination conditions, and that the shape of each point can be solved by the observed variation in shading of the images. 

Being faced with the variations in the wild photos, it is difficult to define the "shape" of a person's face, for which the normal information is also undetermined. Our solution is to solve a face shape that is locally similar to as many photos as possible from the normalized wild images as done in the paper~\cite{30}. Firstly, $n$ images are normalized and clipped to frontal-posed with $p$ pixels, and input into $M_{n\times p}$. Secondly, we estimate the initial shape $\tilde S$ and lighting $\tilde L$ by factorizing $M=LS$ via SVD\cite{42}. $\tilde L=U\sqrt D$ and $\tilde S=\sqrt D V^T$, where $M = UD {V^T}$.

To approach the true normal information, we estimate the shape $S$ and ambiguity $A$ by following the approach in paper ~\cite{30}. Firstly, we update $\tilde S$ firstly by minimizing ${\min _{\tilde S}}\left\| {M - \tilde L \tilde S} \right\| + {S^T}GS$ to get the initial normal $S$, where $G=diag(-1,1,1,1)$, with initial $\tilde L$. Secondly, we solve ambiguity $A$ by  minimizing  ${\min _A}{\left\| {S^t - A \tilde S} \right\|^2}$, where $S^t$ is the normal shape of a template and it is consistent with $S$ at level of pixel position. At last, the surface normals are obtained by $S=A\tilde S$.

Particularly, the consistent normal shape $S^t$ are from the normals of template  $\frac{{\textbf{\emph{F}}_u '  \times \textbf{\emph{F}}_v '}}{{\left\| {\textbf{\emph{F}}_u ' \times \textbf{\emph{F}}_v '} \right\|}} =  \Lambda \cdot \textbf{\emph{F}}_u ' \otimes \textbf{\emph{F}}_v '$, where symbol $\otimes$ is compound operator of cross-product $\times$ in Equ.\ref{equ:03}. 
    
    The definition of $\otimes$ operation is 
   $\textbf{\textit{w}} \otimes \textbf{\textit{v}} = \left[ {\begin{array}{*{20}{c}}
{{\textbf{\textit{w}}_1} \times {\textbf{\textit{v}}_1}}\\
{{\textbf{\textit{w}}_2} \times {\textbf{\textit{v}}_2}}\\
 \vdots \\
{{\textbf{\textit{w}}_p} \times {\textbf{\textit{v}}_p}}
\end{array}} \right]$,
where $\textbf{\textit{w}}$ and $\textbf{\textit{v}}$ are $3p\times 1$ vectors containing $p$ normals.
    
    Therefore, the vecorization of $S^t$ is 
\[\textbf{\textit{h}}_t= \Lambda |_{\textbf{\textit{b}}_t}\cdot (\textbf{\emph{F}}_u ' \otimes \textbf{\emph{F}}_v ')^{(sn)}
=\Lambda |_{\textbf{\textit{b}}_t}\cdot (( \textbf{\emph{T}}_1^{(sn)}  \textbf{\emph{b}}_t) \otimes (\textbf{\emph{T}}_2^{(sn)}  \textbf{\emph{b}}_t))\], where vector $\textbf{\textit{h}}_t$ is $3p \times 1$, and $(*)^{(sn)}$ is a selection operator that selects $3p$ rows of 1st-order coefficients corresponding to $p$ normals consistent with $S$; and $\Lambda$, whose value depends on $\textbf{\textit{b}}_t$ is a $3p\times 3p$ diagonal matrix that stores $3p$ reciprocals of lengths of the $p$ selected normals; and $\textbf{\emph{b}}_t$ represent the control points of B-spline template face. 

Similarly, the obtained $S$ can also be vectorized as $\textbf{\textit{h}}$ that is $3p\times 1$. Then $\textbf{\textit{h}}$ can be used to optimize control points $\textbf{\textit{b}}$ for B-spline objective face conversely, which will be introduced in the following subsection.

\subsubsection{Local B-spline optimization}

According to the relation of normals vector $\textbf{\textit{h}}$ and B-spline control points \textbf{\textit{b}}, B-spline face can be solved by
\[\mathop {\min }\limits_\textbf{\textit{b}}  \textbf{O}_1=\mathop {\min }\limits_\textbf{\textit{b}}  \left\| {\textbf{\textit{h}} - \Lambda |_{\textbf{\textit{b}}}  \cdot ((\textbf{\textit{T}}_1^{(sn)}\textbf{\textit{b}}) \otimes ((\textbf{\textit{T}}_2^{(sn)}\textbf{\textit{b}}))} \right\|,\]
which is also referred to as 1st-order optimization.

However, there exists two issues: 1) the low-dimension $\textbf{\textit{h}}$ may not guarantee an unique solution of high-dimension $\textbf{\textit{b}}$; and 2) the system is not simply linear, which is difficult to be solved. 

Therefore, a frontal constraint with $\textbf{\textit{b}}_t$ is applied to make a unique solution; And a strategy of approximating to linearization is also proposed to make a linear solution.

\noindent \textbf{\textit{Frontal Constraint.}}  
As the normals are estimated actually from frontal-face images, a frontal constraint will enhance an effective solution of 1st-order optimization surely by 
\[{\textbf{O}_2} = \left\| {\textbf{\textit{T}}^{(sxy)} \textbf{\textit{b}} - \textbf{\textit{T}}^{(sxy)} {\textbf{\textit{b}}_t}} \right\|<\epsilon ,\]
where matrix $\textbf{\textit{T}}$ is the stacked coefficients of control points, and $(*)^{(sxy)}$ is a selection operator that sets the coefficients corresponding to z- components of control points to zeros, retaining the coefficients corresponding to x- and y- components. 

This constraint is also referred as 0th-order regularization for local B-spline optimization as following,
\[\mathop {\min }\limits_\textbf{\textit{b}} {\textbf{\textit{O}}_1}, \ \ \ with \ \ \ {\textbf{\textit{O}}_2}<\epsilon.\]

Particularly, the item $\textbf{\textit{O}}_2$ is a linear form, but the first item $\textbf{\textit{O}}_1$ is not a simple linear form. Therefore, an approximating to linearization is proposed.

\noindent \textbf{\textit{Approximating to Linearization.}}
According to the characteristics of the cross-product “$\otimes$”, the first item $\textbf{\textit{O}}_1$ can be rewritten as a linear-like formulation: 
\[ \left\| {\textbf{\textit{h}} - {\boldsymbol L} {|_\textbf{\textit{b}}} \cdot \textbf{\textit{b}}} \right\| \ \ or \ \ \left\| {\textbf{\textit{h}} - {\boldsymbol R} {|_\textbf{\textit{b}}} \cdot \textbf{\textit{b}}} \right\|,\]
where \[{\boldsymbol L}{|_\textbf{\textit{b}}} = \Lambda {|_\textbf{\textit{b}}} \cdot {\left[ {\textbf{\textit{T}}_1^{(sn)}\textbf{\textit{b}}} \right]_ \otimes } \cdot \textbf{\textit{T}}_2^{(sn)}\] and \[\textbf{\textit{R}}{|_\textbf{\textit{b}}} =  - \Lambda {|_\textbf{\textit{b}}} \cdot {\left[ {\textbf{\textit{T}}_2^{(sn)}\textbf{\textit{b}}} \right]_ \otimes } \cdot \textbf{\textit{T}}_1^{(sn)}.\] 

Particularly, the operation $[*]_\otimes$ makes vector $\textbf{\textit{w}}_{3p\times 1}$ a $3p\times 3p$ matrix $\left[ \textbf{\textit{w}}\right]_\otimes=diag(\left[ {{\textbf{\textit{w}}_1}} \right]_ \times ,\left[ {{\textbf{\textit{w}}_2}} \right]_ \times ,...,\left[ {{\textbf{\textit{w}}_p}} \right]_ \times )$, where $\textbf{\textit{w}}=\left[ \textbf{\textit{w}}_1;
\textbf{\textit{w}}_2;...;\textbf{\textit{w}}_p \right]$, and \\ \[ \left[ {{\textbf{\textit{w}}_i}} \right]_ \times=\left[ {\begin{array}{*{20}{c}}
0&{ - {\textbf{\textit{w}}_i^z}}&{{\textbf{\textit{w}}_i^y}}\\
{{\textbf{\textit{w}}_i^z}}&0&{ - {\textbf{\textit{w}}_i^x}}\\
{ - {\textbf{\textit{w}}_i^y}}&{{\textbf{\textit{w}}_i^x}}&0
\end{array}} \right], i=1,2,...p.\]

If $\textbf{\textit{b}}$ is a known parameter, e.g as $\textbf{\textit{b}}_0$, for ${\boldsymbol L}|_\textbf{\textit{b}}$ and ${\boldsymbol R}|_\textbf{\textit{b}}$, the minimization of both
$ \left\| {\textbf{\textit{h}} - {\boldsymbol L} {|_\textbf{\textit{b}}}_0 \cdot \textbf{\textit{b}}} \right\| \ \ and \ \ \left\| {\textbf{\textit{h}} - {\boldsymbol R} {|_\textbf{\textit{b}}}_0 \cdot \textbf{\textit{b}}} \right\| $ will be a linear system.

In fact, formulation $\left\| {\textbf{\textit{h}} - {\boldsymbol L} {|_\textbf{\textit{b}}}_0 \cdot \textbf{\textit{b}}} \right\|$ is used to optimize the control points in parameter space of $v$ by fixing $u$, while $\left\| {\textbf{\textit{h}} - {\boldsymbol R} {|_\textbf{\textit{b}}}_0 \cdot \textbf{\textit{b}}} \right\|$ is used to optimize the control points in parameter space of $u$ by fixing $v$. 

\begin{figure}[t]
\begin{center}
\includegraphics[width=1.0\linewidth]{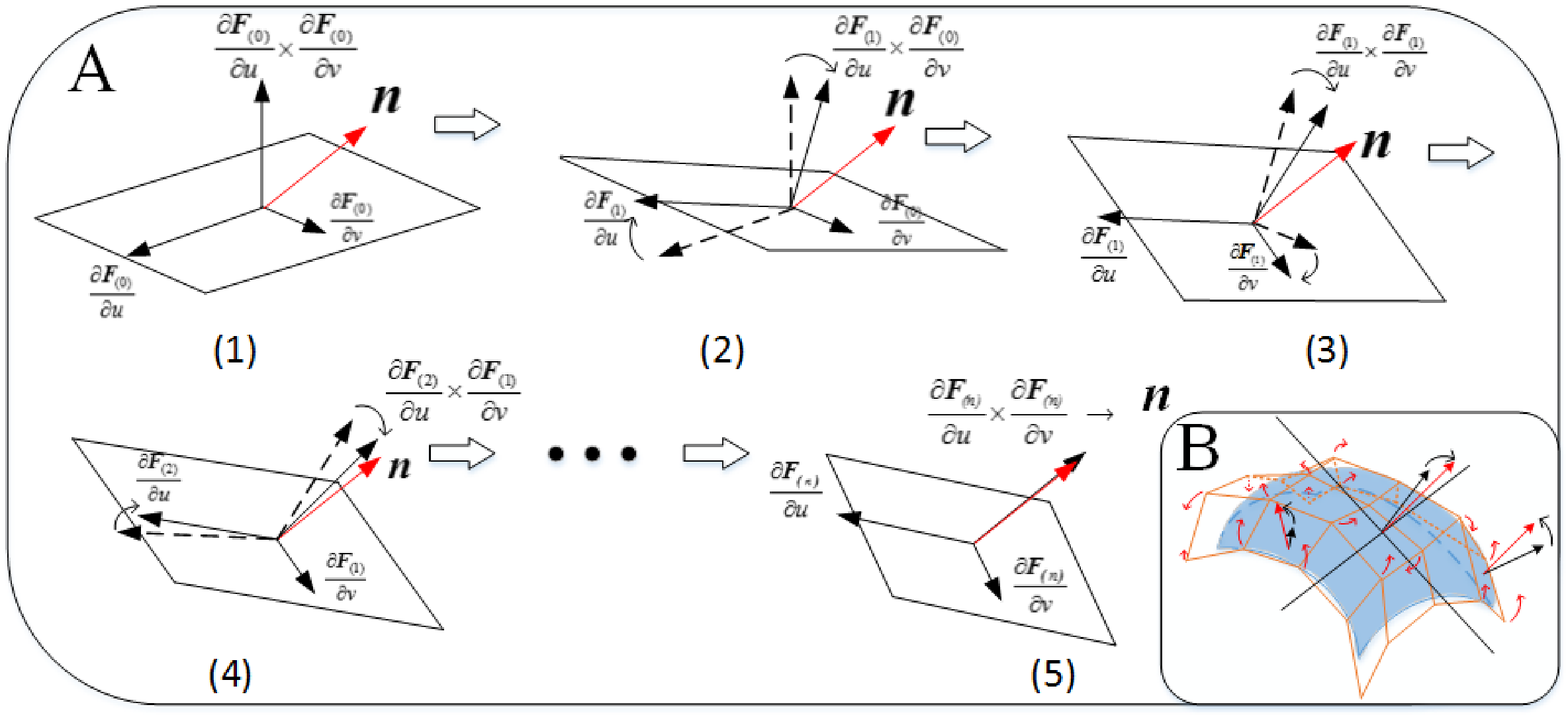}
\end{center}
   \caption{ Iterative adjustment on two partial derivatives: Process (1) to (2) adjusts $\frac{{\partial \textbf{\textit{F}}}}{{\partial u}}$ by fixing $\frac{{\partial \textbf{\textit{F}}}}{{\partial v}}$, and process (3) to (4) adjusts $\frac{{\partial \textbf{\textit{F}}}}{{\partial v}}$ by fixing $\frac{{\partial \textbf{\textit{F}}}}{{\partial u}}$, ... until that $\frac{{\partial \textbf{\textit{F}}}}{{\partial u}} \times\frac{{\partial \textbf{\textit{F}}}}{{\partial v}}$ is infinitely close to objective $\textbf{\textit{n}}$; Process A implements a practically and iteratively linear handle for B-spline surface adjustment in B.}
\label{fig:meth_normal}
\end{figure}

A practical skill is to optimize the control points on the two parameter spaces alternately and iteratively. The two iteration items for $\textbf{O}_1$ can be rewritten as
\[\left\| {\textbf{\textit{h}} - {\boldsymbol L}{|_{{\textbf{\textit{b}}_0}}} \cdot \textbf{\textit{b}}} \right\| + \lambda \left\| {\textbf{\textit{T}}_1^{(sn)} \cdot \textbf{\textit{b}} - \textbf{\textit{T}}_1^{(sn)} \cdot {\textbf{\textit{b}}_t}} \right\|\]
and
\[\left\| {\textbf{\textit{h}} - {\boldsymbol R}{|_{{\textbf{\textit{b}}_0}}} \cdot \textbf{\textit{b}}} \right\| + \lambda \left\| {\textbf{\textit{T}}_2^{(sn)} \cdot \textbf{\textit{b}} - \textbf{\textit{T}}_2^{(sn)} \cdot {\textbf{\textit{b}}_t}} \right\|.\]
As shown in Fig.\ref{fig:meth_normal}, two partial derivatives $\frac{{\partial \textbf{\textit{F}}}}{{\partial v}}$ and $\frac{{\partial \textbf{\textit{F}}}}{{\partial u}}$ at $(u,v)$ are updated until $\frac{{\partial \textbf{\textit{F}}}}{{\partial v}} \times \frac{{\partial \textbf{\textit{F}}}}{{\partial u}}$ converges to $\textbf{\textit{n}}$. Finally, the whole surface approaches to the optimal.

By integrating with $\textbf{O}_2$,  the final formulation for local optimization is the two iteration items as following:
\begin{equation}\footnotesize
\mathop {\min }\limits_\textbf{\textit{b}} \left\| {\left[ {\begin{array}{*{20}{c}}
\textbf{\textit{h}}\\
{{\textbf{\textit{T}}^{(sxy)}}{\textbf{\textit{b}}_t}}
\end{array}} \right] - \left[ {\begin{array}{*{20}{c}}
{{\boldsymbol L}{|_{{\textbf{\textit{b}}_0}}}}\\
{{\textbf{\textit{T}}^{(sxy)}}}
\end{array}} \right]\textbf{\textit{b}}} \right\| + \lambda \left\| {\textbf{\textit{T}}_1^{(sn)}\textbf{\textit{b}} - \textbf{\textit{T}}_1^{(sn)}{\textbf{\textit{b}}_t}} \right\|
\label{equ:06}
\end{equation}
and
\begin{equation}\footnotesize
\mathop {\min }\limits_\textbf{\textit{b}} \left\| {\left[ {\begin{array}{*{20}{c}}
\textbf{\textit{h}}\\
{{\textbf{\textit{T}}^{(sxy)}}{\textbf{\textit{b}}_t}}
\end{array}} \right] - \left[ {\begin{array}{*{20}{c}}
{{\boldsymbol R}{|_{{b_0}}}}\\
{{\textbf{\textit{T}}^{(sxy)}}}
\end{array}} \right]\textbf{\textit{b}}} \right\| + \lambda \left\| {\textbf{\textit{T}}_2^{(sn)}\textbf{\textit{b}} - \textbf{\textit{T}}_2^{(sn)}{\textbf{\textit{b}}_t}} \right\|
\label{equ:07}
\end{equation}

The $\textbf{\textit{b}}_0$ is inited by value of $\textbf{\textit{b}}_t$. Then we can solve $\textbf{\textit{b}}$ and update $\textbf{\textit{b}}_0$ orderly by Minimization (\ref{equ:06}) and  (\ref{equ:07}) iteratively until convergence. Please refer to  line ~\ref{code:shading_begin}-~\ref{code:shading_end} in Algorithm~\ref{alg:Framework} for details.

\subsection{Global B-spline Shape From Motion}
\label{sec:2.3}
The motion informations across images cover the geometry projections of feature points of a 3D object on 2D projection planes, which also can be used to optimize B-spline control points of a B-spline face. What's important, the motion informations feature their sparsity and globality. Firstly, we use SVD method to re-estimate the sparse motion information by assuming that face images are captured from a B-spline shape by a camera model of a scaled orthographic projection. Secondly, the re-estimated motion information is used to optimize the shape of B-spline face globally. The two steps also are conducted iteratively.

\subsubsection{Re-estimating the Motion}
According to relation between facial points $q$ on image and $Q$ on B-spline face: $q = sRQ + t$, pose parameters $s$,$R$ and $t$ for each image are recovered based on the linear transformation and SVD method ~\cite{30}.  

Note a vector $\textbf{\textit{f}}_{2nf\times 1}$ stacking  $f$ feature points $(q-t)$ of $n$ images as the the re-estimated 2D positions, and a projection matrix $\textbf{\textit{P}}_{2nf\times 3f}$ stacking $n$ views of parameters $sR$. 
\subsubsection{Global B-spline Warping}

According to the projection relation between 2D feature points and and 3D feature points, warping is based on an optimized B-spline face $\textbf{\textit{b}}_0$ that contains local information of objective face:
\[\mathop {\min }\limits_\textbf{\textit{b}} \left\| {\textbf{\textit{f}} - \textbf{\textit{P}} \cdot {\textbf{\textit{T}}^{(sf)}}\textbf{\textit{b}}} \right\| +\zeta\cdot l(\textbf{\textit{b}},\textbf{\textit{b}}_0),\]
where $(*)^{(sf)}$ is a selection operator that selects $3f$ rows of coefficients corresponding to $f$ feature points, and $\textbf{\textit{T}}^{(sf)}$ is a $3f\times 3MN$ matrix that stacks $\textbf{\textit{T}}|_{u(i),v(i)}, i=1,2,...,f$. Regularity item $l(\textbf{\textit{b}},\textbf{\textit{b}}_0)$ is the distance of local information between faces $\textbf{\textit{b}}$ and $\textbf{\textit{b}}_0$, and $\zeta$ is a weight factor. The first item is also referred to 0st-order item.

To eliminate affect of geometric rotation brought by 0st-order warping which changes the local normals greatly, and to make sure a smoothness of shape, we regularize the 0st-order item by using 2nd-order derivative information. 

\noindent \textbf{\textit{2nd-order Regularization.}}  According to Equ.\ref{equ:02} that represents 1st-order partial derivative as $\textbf{\textit{F}}{'_u} = {\textbf{\textit{T}}_1} \cdot \textbf{\textit{b}}$, the 2nd-order partial derivative can be introduced as
\[\left\{ {\begin{array}{*{20}{c}}
{\textbf{\textit{F}}''_{uu} = {\textbf{\textit{T}}_{11}} \cdot \textbf{\textit{b}}}\\
{\textbf{\textit{F}}''_{vv} = {\textbf{\textit{T}}_{22}} \cdot \textbf{\textit{b}}}\\
{\textbf{\textit{F}}''_{uv} = {\textbf{\textit{T}}_{12}} \cdot \textbf{\textit{b}}}
\end{array}} \right.,\]
in a similar way. The regularization item representing the 2nd-order approximation between template and objective face is given as:
\[ l(\textbf{\textit{b}},\textbf{\textit{b}}_0) = \left\| {\left[ {\begin{array}{*{20}{c}}
{{\textbf{\textit{T}}_{11}}}\\
{{\textbf{\textit{T}}_{22}}}\\
{{\textbf{\textit{T}}_{12}}}
\end{array}} \right]\textbf{\textit{b}} - \left[ {\begin{array}{*{20}{c}}
{{\textbf{\textit{T}}_{11}}}\\
{{\textbf{\textit{T}}_{22}}}\\
{{\textbf{\textit{T}}_{12}}}
\end{array}} \right]{\textbf{\textit{b}}_0}} \right\|,\]
which is also referred to as 2nd-order regularization.

With the 0th- and 2nd- order information, a globally warping of B-spline face can be formulated as
{\small
\begin{equation}
\mathop {\min }\limits_\textbf{\textit{b}} \left\| {\textbf{\textit{f}} - \textbf{\textit{P}} \cdot {\textbf{\textit{T}}^{(sf)}}\textbf{\textit{b}}} \right\| +\zeta\left\| {\left[ {\begin{array}{*{20}{c}}
{{\textbf{\textit{T}}_{11}}}\\
{{\textbf{\textit{T}}_{22}}}\\
{{\textbf{\textit{T}}_{12}}}
\end{array}} \right]\textbf{\textit{b}} - \left[ {\begin{array}{*{20}{c}}
{{\textbf{\textit{T}}_{11}}}\\
{{\textbf{\textit{T}}_{22}}}\\
{{\textbf{\textit{T}}_{12}}}
\end{array}} \right]{\textbf{\textit{b}}_0}} \right\|
\label{equ:08}
\end{equation}}

Motion re-estimation and B-spline warping are also carried on iteratively, during which 2D position $\textbf{\textit{f}}$, projection $\textbf{\textit{P}}$, objective $\textbf{\textit{b}}$ and template $\textbf{\textit{b}}_t$ are updated in turn, until that template $\textbf{\textit{b}}_t$ is converged. Please refer to  line ~\ref{code:motion_begin}-\ref{code:motion_end} in Algorithm~\ref{alg:Framework} for details.

\subsection{Algorithm}
\label{sec:2.4}

\begin{algorithm}[htb]
\caption{IMLS Algorithm for B-spline face optimization from Shading \& Motion}
\label{alg:Framework}
\begin{algorithmic}[1]
\REQUIRE A set of face images, B-spline template face $\textbf{\textit{b}}_t$, and $uv$ parameters $(u_i,v_i)$, $i=1,2,...,f$ for its feature points. 

\STATE Detect the feature points of images
\WHILE {$\textbf{\textit{b}}$ is not converged}   \label{code:outloop_begin} 
\STATE Let $\textbf{\textit{b}}_0=\textbf{\textit{b}}_t$. {\footnotesize{//  \textit{\texttt{Inited with template}}}}\\
{\footnotesize{//  \textit{\texttt{Compute local B-spline from shading}}}}
\WHILE {$\textbf{\textit{b}}_0$ is not converged} \label{code:shading_begin}
    \STATE Conduct photometric stereo to obtain normal vector \textbf{\textit{h}} using template $\textbf{\textit{b}}_0$.
    \\ {\footnotesize{//  \textit{\texttt{Optimize shape in parameter space of $v$}}}}
    \STATE With known $\textbf{\textit{b}}_0$, solve Minimization (\ref{equ:06}) by
    {\scriptsize
    \[\textbf{\textit{b}}^{(1)} = {({{\boldsymbol G}^T}{\boldsymbol G} + \lambda {\left[ {\textbf{\textit{T}}_1^{(sn)}} \right]^T}\textbf{\textit{T}}_1^{(sn)})^{ - 1}} \cdot ({{\boldsymbol G}^T}\textbf{\textit{h}} + \lambda {\left[ {\textbf{\textit{T}}_1^{(sn)}} \right]^T} {\textbf{\textit{T}}_1^{(sn)}} {\textbf{\textit{b}}_0}),\]
    where ${\boldsymbol G} = \left[ {\begin{array}{*{20}{c}}
{{\boldsymbol L}{|_{{\textbf{\textit{b}}_0}}}}\\
{{\textbf{\textit{T}}^{(sxy)}}}
\end{array}} \right] $.} \label{code:solution6}
\\ {\footnotesize{//  \textit{\texttt{Optimize shape in parameter space of $u$}}}}
    \STATE Let $\textbf{\textit{b}}_0=\textbf{\textit{b}}^{(1)}$, and solve Minimization (\ref{equ:07}) by
    {\scriptsize
    \[\textbf{\textit{b}}^{(2)} = {({{\boldsymbol H}^T}{\boldsymbol H} + \lambda {\left[ {\textbf{\textit{T}}_2^{(sn)}} \right]^T}T_2^{(sn)})^{ - 1}} \cdot ({{\boldsymbol H}^T}\textbf{\textit{h}} + \lambda {\left[ {\textbf{\textit{T}}_2^{(sn)}} \right]^T}{\textbf{\textit{T}}_2^{(sn)}}{\textbf{\textit{b}}_0}),\]
    where ${\boldsymbol H} = \left[ {\begin{array}{*{20}{c}}
{{\boldsymbol R}{|_{{b_0}}}}\\
{{\textbf{\textit{T}}^{(sxy)}}}
\end{array}} \right]$.}  \label{code:solution7}

\STATE Let $\textbf{\textit{b}}_0=\textbf{\textit{b}}^{(2)}$. {\footnotesize{//  \textit{\texttt{Update local shape.}}}}
     \ENDWHILE  \label{code:shading_end}
\\ {\footnotesize{//  \textit{\texttt{Warp B-spline globally via motion}}}}

\WHILE {$\textbf{\textit{b}}_t$ is not converged} \label{code:motion_begin}
    \STATE Re-estimate motion information $\textbf{\textit{f}}$ and $\textbf{\textit{P}}$ using the B-spline template $\textbf{\textit{b}}_t$.
    \STATE With known $\textbf{\textit{b}}_0$, solve Minimization (\ref{equ:08}) by: 
    {\footnotesize
    \[\textbf{\textit{b}} = {({(\textbf{\textit{P}}{\textbf{\textit{T}}^{(sf)}})^T}\textbf{\textit{P}}{\textbf{\textit{T}}^{(sf)}} + \zeta {\boldsymbol {K}^T}\boldsymbol {K})^{ - 1}} \cdot ({(\textbf{\textit{P}}{\textbf{\textit{T}}^{(sf)}})^T}f + \zeta {\boldsymbol {K}^T}\boldsymbol {K}{\textbf{\textit{b}}_0}),\]}
    where ${\boldsymbol K} = \left[ {\begin{array}{*{20}{c}}
{{\textbf{\textit{T}}_{11}}}\\
{{\textbf{\textit{T}}_{22}}}\\
{{\textbf{\textit{T}}_{12}}}
\end{array}} \right]$. \label{code:solution8}
\STATE Let $\textbf{\textit{b}}_t=\textbf{\textit{b}}$. {\footnotesize{//  \textit{\texttt{Update template}}}}
     \ENDWHILE \label{code:motion} \label{code:motion_end}
      \ENDWHILE \label{code:outloop_end} 
\ENSURE ~~\\
Solution of B-spline objective face $\textbf{\textit{b}}$.
\end{algorithmic}
\end{algorithm}

An IMLS (iterative multi-least-square) algorithm is proposed to optimize the B-spline face, seen in Algorithm ~\ref{alg:Framework}. Processes of computing local B-spline shape from shading and warping B-spline globally via motion are carried on iteratively to get the final surface, differing from the work in ~\cite{40} where the uses of shading and motion are in two separate processes. Particularly, Minimization (\ref{equ:06}) and Minimization (\ref{equ:07}) during local optimization, and Minimization (\ref{equ:08}) during global optimization are linear system with closed-from solutions of least square, seen in line~\ref{code:solution6},\ref{code:solution7},\ref{code:solution8} in Algorithm~\ref{alg:Framework}. 

\begin{figure*}[t]
\begin{center}
   \includegraphics[width=1.0\linewidth]{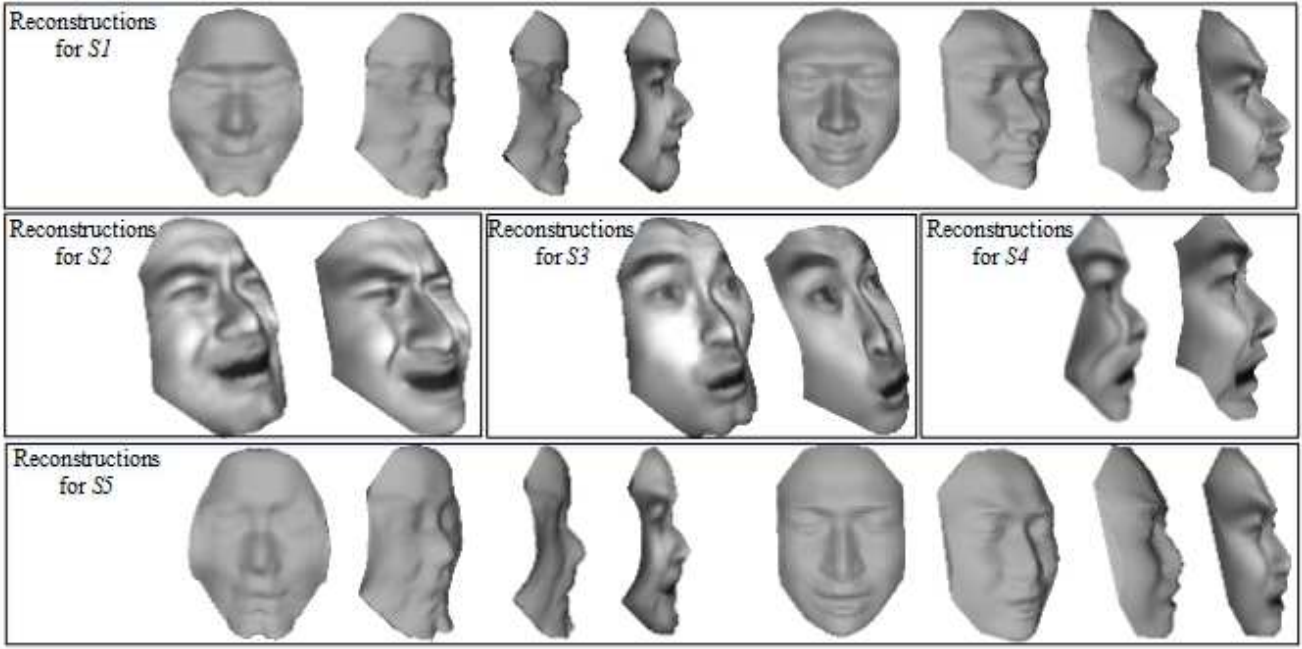}
\end{center}
   \caption{ Visual reconstructions and comparisons for sessions \textit{S1}, \textit{S2}, \textit{S3}, \textit{S4} and \textit{S5}: for each session of reconstructions, the left shows the results by reimplemented method\cite{30}, and the right shows our results.}
\label{fig:exp_result1}
\end{figure*}

\begin{figure}[!]
\begin{center}
   \includegraphics[width=1.0\linewidth]{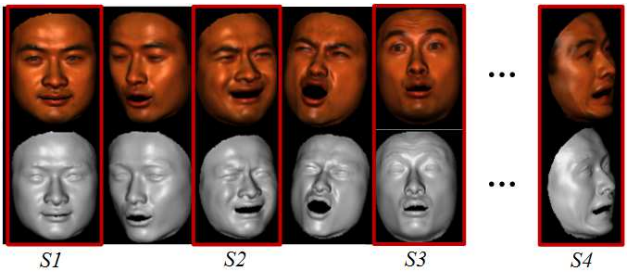}
\end{center}
   \caption{Sample data simulated by the spaces-times faces\cite{3} : images and 3D model with various poses and illuminations are available; samples \textit{S1},\textit{S2},\textit{S3}, and \textit{S4} are used for evaluation.
}
\label{fig:exp_samples}
\end{figure}

\begin{figure*}[t]
\begin{center}
   \includegraphics[width=1.0\linewidth]{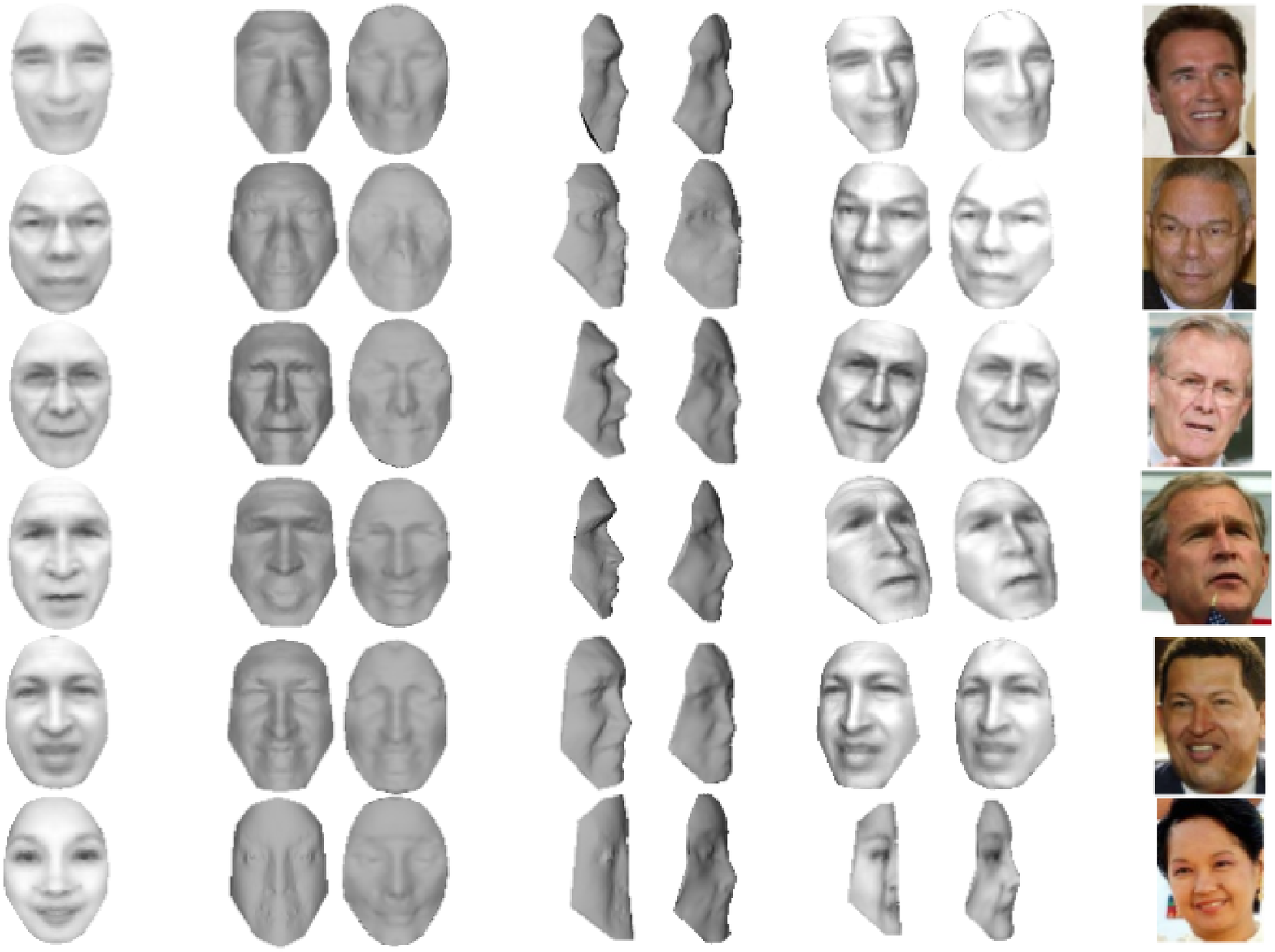}
\end{center}
   \caption{Visual reconstructions and comparisons for Arnold Schwarzenegger, Colin Powell, Donald Rumsfeld, George W.Bush, Hugo Chavez and Gloria Macapagal Arroyo in LFW~\cite{ 41}: 1) Images in column 1 and 8  are mean textures and true images respectively; 2) Columns 2, 4, and 6 show the results by the reimplemented method\cite{30}, and columns 3, 5, and 7 show our results; 3) columns 6 and 7 are rendered with mean texture.
}
\label{fig:exp_result2}
\end{figure*}
\section{Experiment}
　In this section experiments is resented to verify our method of MsSfMS. We first describe the pipeline to prepare a collection of face images of a same person for B-spline face reconstruction. Then quantitative and qualitative comparisons are demonstrated with baseline method on ground truth data~\cite{3} with various expressions, illuminations and poses. Finally, challenging reconstructions are also conducted and compared based on real unconstrained data taken from the Labeled Faces in Wild (LFW)~\cite{41} database.
　
\subsection{Pipeline of experimental data}
\noindent \textbf{Ground truth data with expression}\ \ \ The ground truth data is from the space-times faces~\cite{3}  which contains 3D face models with different expressions. And different poses and illuminations can also be simulated by the spaces-times faces, seen in Fig.\ref{fig:exp_samples}. Images with various poses and illuminations are collected, and feature points manually labeled.  The reconstruction is evaluated by the error to the ground truth model. 

\noindent \textbf{LFW data}\ \ \ Images of each person are collected and input into a facial point detector~\cite{33} that has a similar high performance to human,  to find the 40 facial points shown in Fig.\ref{fig:meth_deriv}.  The initial B-spline template face is computed from a neutral model of space-time faces.

\subsection{Standard images}
We conduct 5 session of reconstructions: the first four is  reconstruction expression \textit{S1}, \textit{S2}, \textit{S3} and \textit{S4} by using their corresponding images, and the firth reconstruction \textit{S5} is based on images with different expressions.  Each session contains 40 images with various illumination and different poses. Reconstruction result is compared with the reimplemented method\cite{30}.

To compare the approaches numerically, we compute the shortest point-to-point distance from ground truth to reconstructed surface. Point clouds are sampled from B-spline face and aligned according to absolute orientation problem. Mean euclidean distance and the root mean square (RMS) of the distances, after normalized by the eye-to-eye distance, are reported in Table.\ref{table1}. Visual comparison is shown in Fig.\ref{fig:exp_result1}. Our results demonstrate promise performance. An important fact is that method\cite{30} cannot make a credible depth information and global shape, but our method solves the problem by local optimization and global warping.

\begin{table}\small
\caption{Distances of the reconstruction to the ground truth}
\begin{center}
\begin{tabular}{r|r|r|r|r|r|r}
\hline
Method & Index & S1 & S2 & S3 & S4 & S5 \\
\hline
Method&                  Mean(\%) & 8.53 & 8.67 & 8.92 & 10.49 & 9.29\\
        \cite{30}         & RMS(\%) & 6.92 & 7.34 & 4.68 & 6.21  & 7.58\\
\hline
Ours                        &Mean(\%) & \textbf{7.09} & \textbf{7.86} & \textbf{7.11} & \textbf{8.92}  & \textbf{8.49}\\
                              & RMS(\%)   & \textbf{4.57} & \textbf{5.61} & \textbf{4.18} & \textbf{5.61}  & \textbf{5.60}\\
\hline
\end{tabular}
\label{table1}
\end{center}
\end{table}
\subsection{Real unconstrained data}
Our method is also tested based on real unconstrained data. Unlike the experiment in paper\cite{30} using hundreds of images, we conduct reconstruction with limited number of images, as so many images are not always available for some task like criminal investigation and small simple size problem. 35 images are collected from LFW for each of six persons (Arnold Schwarzenegger, Colin Powell, Donald Rumsfeld, George W.Bush, Hugo Chavez and Gloria Macapagal Arroyo), covering different pose and expressions. Reconstructions and comparisons are shown in Fig.\ref{fig:exp_result2}. Our method steadily produce good looking model from different viewpoints. Especially for case of Gloria Macapagal Arroyo, while the method\cite{30} is not working, our method generates satisfactory result. It is convinced that the reconstructions express the individual face structure well and fit well with pose-variety photos. 

\section{Conclusions}
We presented an B-spline free-form surface modeling for face reconstruction,by leveraging the spline cues of motion and shading in a set of face images. The method works well for data with different poses and expressions, even for wild cases.  The key contributions are that the proposed $C^2$ B-spline shape model describes the motion and shading geometrically in 0th- and 1st- order derivatives respectively, and that an optimization algorithm is also supplied for reconstruction. In the future, the model also can be extented for other research areas, such as expression tracking and facial animation by adjusting the control points locally, and recognition and verification based on local B-spline shape information.

{\small
\bibliographystyle{ieee}
\bibliography{egbib}
}

\end{document}